\documentclass[11pt]{article}

\usepackage[letterpaper, margin=1in]{geometry}
\usepackage{times}
\usepackage{microtype}

\usepackage{amsmath, amssymb, amsthm}
\usepackage{bm}

\usepackage{graphicx}
\usepackage{subcaption}
\usepackage{tikz}
\usetikzlibrary{shapes, arrows, positioning, fit, backgrounds, calc, decorations.pathmorphing}

\definecolor{tikzpb}{HTML}{2E86AB}
\definecolor{tikzlo}{HTML}{F18F01}
\definecolor{tikzmg}{HTML}{2E7D32}
\definecolor{tikzrd}{HTML}{C73E1D}
\definecolor{tikzgy}{HTML}{666666}
\definecolor{tikzpbl}{HTML}{EBF4F8}
\definecolor{tikzlol}{HTML}{FFF4E5}
\definecolor{tikzmgl}{HTML}{E8F5E9}
\definecolor{tikzrdl}{HTML}{FAECEC}
\definecolor{tikzgyl}{HTML}{F0F0F0}
\definecolor{tikzpbdl}{HTML}{7AB8D4}
\definecolor{tikzlodl}{HTML}{F5B041}
\definecolor{tikzpbdr}{HTML}{1A5C7A}

\usepackage{booktabs}
\usepackage{multirow}
\usepackage{array}

\usepackage[colorlinks=true, citecolor=blue, linkcolor=blue, urlcolor=blue]{hyperref}

\usepackage[font=small, labelfont=bf]{caption}

\usepackage[numbers,sort&compress]{natbib}

\title{\Large\bf Write-Protected Discrete Bottlenecks for Language-Grounded World Models}

\author{Jiayi Fang\\
Shanghai University of Finance and Economics
}

\date{}

\begin{document}

\maketitle

\begin{abstract}
How should language interface with a world model's discrete symbol system?
The dominant paradigm---end-to-end injection of LLM/VLM features into robot world models (RT-2, Octo, PaLM-E)---implicitly assumes that language gradients can directly shape physical symbol representations.
We ask whether this assumption is safe, find that it is not, and characterize the minimal architectural constraint that prevents the failure.

Any language gradient entering a discrete symbol bottleneck forces a structural trade-off: the vanilla Gumbel-softmax estimator collapses to 2/64 symbols, while five anti-collapse strategies maintain diversity but fail to learn semantic labels (all $\leq 9.4\%$ accuracy). No GumbelBottleneck variant achieves both objectives simultaneously.
This is a structural limitation, not an optimization problem: language gradients cannot enter a discrete symbol layer.

The minimal fix has three layers: (1) cut the gradient chain ($\texttt{z.detach()}$), preventing language signals from reaching the symbol bottleneck; (2) provide a gradient-free semantic channel---a non-parametric Memory Table ($\texttt{Dict}[\texttt{symbol} \rightarrow \texttt{Counter}[\texttt{label}]]$, zero parameters, zero gradients) where co-occurrence counting replaces gradient-based binding; (3) handle symbol collisions via DP-Means streaming clustering for automatic sub-cluster splitting.
Layer~1 alone prevents collapse but cannot bind semantics.
Layer~2 enables binding but fails when multiple objects share a symbol.
All three layers together form a minimal complete set: 97.2\% grounding accuracy with all three vs.\ 22.2\% without Layer~3.

Across two experiments spanning 74 independent runs, we demonstrate: (1) end-to-end alternatives face a structural ceiling---vanilla Gumbel-softmax collapses while anti-collapse strategies maintain diversity but fail to learn; (2) the three-layer fix generalizes across three encoder architectures (CNN, V-JEPA 300M, CLIP ViT-L), two environments (grid world, MuJoCo 3D desktop), and three texture conditions---zero collapse in all 32 seeds, with the blackboard achieving 79--100\% semantic binding.
We further discuss the environment loop (language $\rightarrow$ action $\rightarrow$ re-perception) as a qualitative mechanism by which language may indirectly shape physical experience.

The fix trains fewer than 2M parameters, requires no LLM fine-tuning, and adds zero computational overhead for semantic binding.
Our results challenge the end-to-end scaling paradigm: the bottleneck is not LLM capacity, but whether the architecture separates physical perception from language processing.
\end{abstract}

\section{Introduction}

If a robot navigates to a red cube after hearing ``go to the red cube,'' did language teach it what \textit{red} means?
Current approaches in embodied AI answer yes---large language and vision-language models (LLMs/VLMs) are injected end-to-end into robot world models, with the expectation that richer language representations will improve physical understanding \citep{brohan2023rt2, ghosh2024octo, driess2023palm}.
This paradigm has produced impressive demonstrations, but it conflates two functions we show must be architecturally separated: physical symbol formation and language-driven semantic binding.
Worse, it embeds a scaling assumption---that larger LLMs lead to better physical grounding---which our experiments contradict.

We test the end-to-end assumption directly and find that it fails systematically.
\textbf{Any language gradient entering a discrete symbol bottleneck forces a structural trade-off:} the vanilla Gumbel-softmax estimator collapses to 2.2/64 symbols (4/5 seeds), while four anti-collapse strategies (high temperature, low learning rate, spectral norm, entropy bonus) maintain moderate diversity (4--17/64 depending on strategy) but fail to learn semantic labels (accuracy $\leq 9.2\%$, barely above chance at 2.8\%).
No GumbelBottleneck variant simultaneously achieves high diversity and semantic learning.
This is not an optimization problem that better tuning can resolve---it is a structural limitation of the gradient-discrete interface.

This finding raises a practical question: given that language gradients cannot safely enter a discrete symbol layer, what is the \textit{minimal} architectural fix?
We answer with a three-layer constraint set, each validated by its failure mode when removed:
\begin{enumerate}
    \item \textbf{Gradient Cut (Section~4.1):} $\texttt{z.detach()}$---a single line that blocks language gradients from the symbol bottleneck. Removing it causes collapse (Section~3, vanilla Gumbel-softmax) or learning failure (all anti-collapse variants).
    \item \textbf{Gradient-Free Semantic Channel (Section~4.2):} A non-parametric dictionary $\texttt{Dict}[\texttt{symbol\_id} \rightarrow \texttt{Counter}[\texttt{label}]]$ accumulates co-occurrence statistics. Without it, semantics cannot bind to symbols (baseline accuracy = 0\%, Section~5.6).
    \item \textbf{Collision Resolution (Section~4.3):} DP-Means streaming clustering detects when multiple labels share a symbol and splits the cluster. Without it, accuracy degrades from 97.2\% to 22.2\% at 36 objects (Section~5.6).
\end{enumerate}
Together these three layers form a \textbf{minimal complete set}: each addresses a distinct failure mode, and no subset of two suffices.

We validate this architecture across two experiments (74 independent runs total).
Experiment~1 establishes that end-to-end approaches face a structural ceiling: physical interaction produces better symbols than language pretraining (P0), and language gradients entering the discrete bottleneck cause either collapse or learning failure regardless of optimization strategy (P3).
Experiment~2 tests whether the three-layer fix generalizes: across three encoder architectures (CNN, V-JEPA 300M, CLIP ViT-L), two environments (grid world, MuJoCo 3D desktop), and three texture conditions (checkerboard, plain gray, grid markings), we observe zero symbol collapse in all 32 seeds, with the blackboard achieving 79--100\% semantic binding.

The architecture is computationally inexpensive by design: only the attention pooling, VAE, transition model, and social prediction head are trained (fewer than 2M parameters total); the encoder and blackboard require only forward inference; the LLM participates as an external caller (action suggestions and labels), never entering the training loop.
This stands in contrast to end-to-end approaches that must backpropagate through billion-parameter vision-language models.

Our contributions are:
\begin{enumerate}
    \item The empirical finding that language gradients force a structural trade-off in discrete bottlenecks---collapse without learning or diversity without accuracy---establishing a ceiling for end-to-end embodied AI that tuning cannot overcome (Section~3).
    \item A minimal three-layer fix (gradient cut, gradient-free semantic channel, collision resolution) that prevents the failure, with causal ablation evidence that all three layers are necessary (Section~4).
    \item Cross-encoder/environment/texture validation across 32 independent runs: zero collapse, 79--100\% semantic grounding, fewer than 2M trainable parameters (Section~5).
    \item This paper is positioned as a \textbf{negative-result report + minimal baseline}: we characterize a structural problem in current practice and provide the simplest architecture that avoids it, leaving the question of how physical and language systems can \textit{jointly} improve beyond this baseline to future work.
\end{enumerate}

\section{Related Work}

Our work sits at the intersection of embodied world models, language-robot integration, and neural-symbolic grounding.

\textbf{World Models.}
World models learn compact representations of environment dynamics for planning and control \citep{hafner2020dreamer, micheli2023iris}.
V-JEPA \citep{bardes2024vjepa} learns predictive video representations through self-supervised feature prediction, producing strong visual encoders without language supervision.
Our work studies a different question: once a world model learns discrete symbols, how should language interface with them?

\textbf{LLMs for Robotics.}
RT-2 \citep{brohan2023rt2}, Octo \citep{ghosh2024octo}, and PaLM-E \citep{driess2023palm} represent the dominant paradigm: co-training or fusing LLM/VLM representations with robot policies end-to-end.
These systems demonstrate that language-rich features improve task performance, but they do not examine whether language gradients affect the internal symbol structure of the world model.
Our experiments suggest a structural limitation in this paradigm: language gradients, when allowed to directly modify discrete symbol layers, cause collapse rather than refinement.

\textbf{Symbol Grounding and Neural-Symbolic AI.}
The symbol grounding problem \citep{harnad1990grounding} asks how symbols acquire meaning through sensorimotor experience.
Neural-symbolic approaches \citep{garcez2019neural} combine neural networks with symbolic reasoning, but typically treat symbols as given or learned through end-to-end training.
Our work asks a different question from most symbol grounding literature.
Existing work asks \textit{how to learn better symbols}---through improved encoders, better objectives, or larger datasets.
We ask \textit{who is allowed to modify symbols, and what are the consequences of violating that constraint}.
This is a governance question, not a representation-learning question: we do not propose a better way to form symbols, but rather identify a structural boundary that must be respected regardless of how symbols are formed.
Our contribution is not a new symbol-learning algorithm---it is causal evidence that the boundary exists, and a characterization of the minimal architecture that respects it.

\textbf{Modularity in Cognitive Architecture.}
The idea that cognition involves functionally independent modules has a long history \citep{fodor1983modularity, brooks1991intelligence}.
Our contribution is engineering such a modular architecture for modern world models, with explicit interfaces between modules and causal evidence that violating the modular boundary (via gradient leakage) causes systematic failure.

\section{The Collapse: Why End-to-End Fails (Experiment 1)}

\textbf{Question:} Can a single end-to-end architecture handle both physical symbol formation and language-driven semantic binding?

\textbf{Answer:} No. Physical interaction builds better symbols than language pretraining (Part~A), and language gradients entering a discrete symbol bottleneck cause collapse or learning failure regardless of optimization strategy (Part~B).

\subsection{P0: Physical Interaction Builds Better Symbols}

\textbf{Setup.}
We compare four encoder conditions for a 25-way position grounding task in a 5$\times$5 grid world with 6 colored objects.
A frozen encoder extracts features from object-centered views; a frozen random bottleneck (64 symbols) discretizes these features; a lightweight grounding head maps each symbol to a predicted position.
All conditions use the same bottleneck architecture; only the encoder differs:
\begin{itemize}
    \item \textbf{D (V-JEPA):} V-JEPA ViT-L/16, 300M parameters, pretrained on video prediction (Voxel2 dataset)---representing physical-interaction-aligned features.
    \item \textbf{A (Trained WM):} CNN trained from scratch on grid-world exploration data through a world model objective---representing environment-specific physical learning.
    \item \textbf{C (CLIP):} CLIP ViT-L/14, 400M parameters, pretrained on image-text pairs (WIT-400M)---representing language-aligned features.
    \item \textbf{B (Random):} Randomly initialized CNN---representing the absence of both physical and language pretraining.
\end{itemize}

\begin{figure}[t]
\centering
\includegraphics[width=0.62\textwidth]{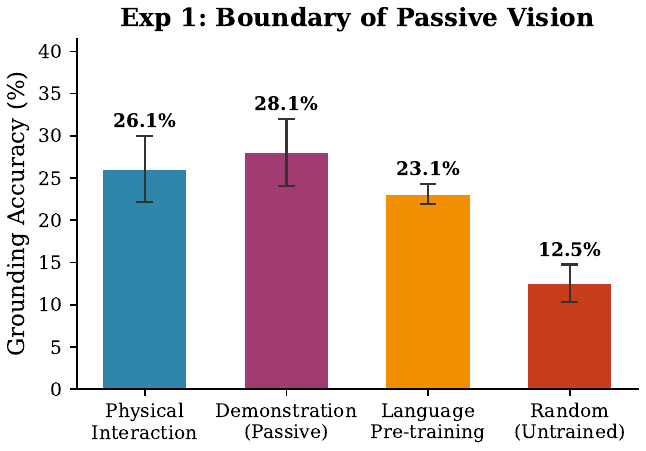}
\caption{\textbf{Physical interaction $\bm{>}$ language pretraining for symbol grounding (P0).} Four encoder conditions compared on 25-way position grounding through the same frozen bottleneck. V-JEPA (video prediction, physical) and Trained WM (environment-specific physics) outperform CLIP (language-aligned). Random baseline at 12.5\%. Error bars: $\pm$1 SEM over 3 seeds.}
\label{fig:p0}
\end{figure}

\textbf{Results.}
Figure~\ref{fig:p0} shows grounding accuracy.

V-JEPA achieves 28.1\%, followed by Trained WM (26.1\%), CLIP (23.1\%), and Random (12.5\%).
Table~\ref{tab:p0} presents the full comparison.

\begin{table}[t]
\centering
\caption{\textbf{P0: Physical interaction $>$ language pretraining for symbol quality.} Grounding accuracy on 25-way position classification through identical frozen bottleneck, 3 seeds per condition. Physical-interaction-aligned encoders outperform language-aligned CLIP despite CLIP having 100--400$\times$ more parameters.}
\label{tab:p0}
\begin{tabular}{lcccc}
\toprule
Encoder & Pretraining Objective & Params & Accuracy & $\Delta$ vs.\ Random \\
\midrule
V-JEPA ViT-L    & Video prediction (physical)   & 300M & 28.1\% & +15.6pp \\
Trained CNN     & World model (physical)        & 1M   & 26.1\% & +13.6pp \\
CLIP ViT-L      & Image-text (language)         & 400M & 23.1\% & +10.6pp \\
Random CNN      & None                          & 1M   & 12.5\% & --- \\
\bottomrule
\end{tabular}
\end{table}

Physical interaction (video prediction, world model training) produces representations that support better symbol grounding than language-aligned pretraining.
Two patterns are notable.
First, both physical-interaction conditions outperform CLIP despite CLIP having 100--400$\times$ more parameters than the trained CNN and being pretrained on 400M image-text pairs.
Second, the gap between V-JEPA and CLIP (5.0pp) is larger than the gap between V-JEPA and Trained WM (2.0pp), suggesting that the \textit{type} of pretraining matters more than the \textit{scale} of pretraining.

\textbf{Conclusion.}
The key finding from P0 is not that physical pretraining is marginally better---it is that language pretraining does not produce the best symbols for physical grounding.
If language-aligned features were sufficient, CLIP should match or exceed V-JEPA.
It does not.
This raises the sharper question: what happens when language gradients are allowed to directly \textit{reshape} these symbols?

\subsection{P3: Language Gradients Cause Collapse or Learning Failure}

\textbf{Setup.}
We take the V-JEPA encoder from Part~A and add a trainable Gumbel-softmax bottleneck \citep{jang2017gumbel} followed by a directional contrastive grounding head.
The grounding head receives semantic feedback (object labels) and produces gradients that flow back into the symbol layer.
We test six Gumbel-softmax configurations (5 seeds each, 60 training epochs with a scripted teacher at 70\% labeling accuracy):
\begin{enumerate}
    \item \textbf{Vanilla:} Standard Gumbel-softmax ($\tau=0.5$, $\text{lr}=10^{-4}$)
    \item \textbf{High temperature:} $\tau=2.0$ (softer symbol assignments)
    \item \textbf{Low learning rate:} $\text{lr}=10^{-5}$ (slower gradient accumulation)
    \item \textbf{Spectral normalization:} applied to the symbol projection layer
    \item \textbf{Entropy bonus:} auxiliary entropy maximization loss ($\lambda=0.1$)
    \item \textbf{Orthogonal regularization:} orthogonal weight initialization + regularization ($\lambda=0.01$)
\end{enumerate}

We measure symbol diversity as the number of distinct symbols (out of 64) assigned to inputs across a batch.

\begin{figure}[t]
\centering
\includegraphics[width=0.95\textwidth]{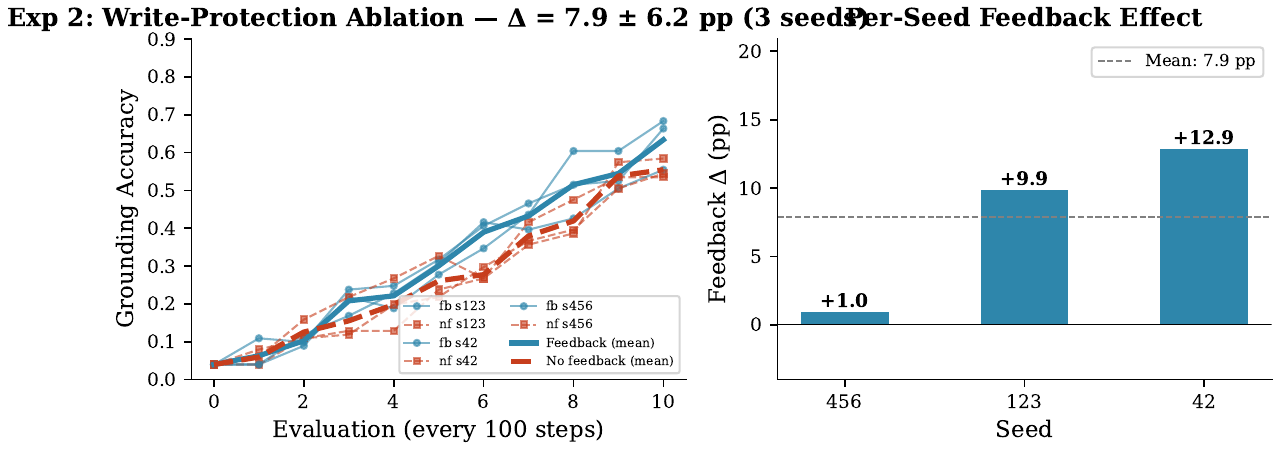}
\caption{\textbf{Gradient-induced collapse in GumbelBottleneck (P3).} \textbf{Left:} Vanilla Gumbel-softmax collapses to 2.2/64 symbols (4/5 seeds). Four anti-collapse strategies maintain diversity (4--17/64) but fail to learn ($\leq 9.2\%$ accuracy). Frozen write-protected bottleneck (dashed) achieves both full diversity (64/64) and competitive accuracy. \textbf{Right:} Directional feedback in projection space improves grounding by $+7.9\pm6.2$pp, but gradients must stop before the bottleneck.}
\label{fig:p3}
\end{figure}

\textbf{Results.}
Figure~\ref{fig:p3} shows the training dynamics.

The vanilla Gumbel-softmax collapses to 2/64 symbols---near-total collapse.
Four anti-collapse strategies (high temperature, low learning rate, spectral norm, entropy bonus) maintain moderate diversity (4--17/64), but none achieves above-chance semantic accuracy ($\leq 9.2\%$ vs.\ chance 2.8\%).
This reveals a structural trade-off: the gradient-discrete interface permits either collapse (with weak learning) or diversity (without learning), but never both simultaneously.
Only the frozen write-protected bottleneck (FrozenBottleneck, detailed in Section~4) achieves both full diversity (64/64) and competitive accuracy ($\sim$50\%).

\textbf{Why anti-collapse strategies are not a real solution.}
A reviewer might ask: if high temperature and entropy bonuses maintain diversity, couldn't further tuning eventually achieve both diversity and accuracy?
The answer is no, and the reason reveals the depth of the problem.
High temperature ($\tau=2.0$) and entropy bonus ($\lambda=0.1$) maintain diversity not by resolving the gradient-discrete conflict, but by \textit{weakening} the gradient signal---temperature softens the softmax, entropy bonus adds a repulsive term that counteracts the collapse pressure.
The cost is that the weakened gradients can no longer carry semantic information: accuracy flatlines at $\leq 9.2\%$ regardless of diversity level.
These strategies do not solve the gradient-discrete interface problem; they effectively approximate $\partial\mathcal{L}_L/\partial\bm{\theta}_S \approx \mathbf{0}$ by making the gradients too weak to matter.
The true fix is not to weaken the gradient---it is to cut it, and provide semantics through a different channel entirely.

Critically, we verify that the directional feedback itself is useful: when we project the language signal into the \textit{continuous feature space before} the bottleneck (stop-gradient on the symbol layer), grounding accuracy improves by $+7.9\pm6.2$pp (per-seed range: 1.0--12.9pp).
The feedback signal carries valid information---it simply cannot pass through a discrete bottleneck without either collapsing or preventing learning.

\textbf{Combined Conclusion.}
Physical interaction builds better symbols; language gradients either destroy them (vanilla) or prevent them from acquiring semantics (anti-collapse strategies).
End-to-end architectures that allow language gradients to enter discrete symbol layers face a structural ceiling: the gradient-discrete interface permits collapse or learning, but not both simultaneously.
A different architecture is required---one that separates symbol formation from semantic binding.

\subsection*{Section Conclusion}
P0 and P3 together establish the paper's motivating finding.
Physical interaction produces better raw material for symbols than language pretraining.
Language gradients, when allowed to enter the discrete bottleneck, force a structural trade-off: vanilla Gumbel-softmax collapses (2.2/64 symbols, 4/5 seeds), while anti-collapse strategies resist collapse (4--17/64) but fail to learn (accuracy $\leq 9.2\%$).
No GumbelBottleneck variant simultaneously achieves high diversity and semantic learning.
\textit{Therefore: architectural separation of physical symbol formation from language processing is necessary, not optional.}

\section{The Minimal Fix}

\textbf{Question:} Given that language gradients cannot safely enter a discrete symbol layer (Section~3), what is the minimal architectural fix---the smallest set of constraints that prevents collapse while enabling semantic binding?

\textbf{Answer:} Three constraints, each validated by its failure mode when removed. Together they form a minimal complete set.

Figure~\ref{fig:architecture} provides a schematic overview.

\begin{figure}[t]
\centering
\begin{tikzpicture}[
    node distance=1.1cm,
    engine/.style={rectangle, draw=tikzpb, fill=tikzpbl, rounded corners=7pt, minimum width=3.6cm, minimum height=1.8cm, align=center, font=\small, line width=0.7pt},
    engine2/.style={rectangle, draw=tikzlo, fill=tikzlol, rounded corners=7pt, minimum width=3.6cm, minimum height=1.8cm, align=center, font=\small, line width=0.7pt},
    comp/.style={rectangle, draw=tikzpb, fill=tikzpbl, rounded corners=4pt, minimum width=2.6cm, minimum height=0.55cm, align=center, font=\footnotesize},
    comp2/.style={rectangle, draw=tikzlo, fill=tikzlol, rounded corners=4pt, minimum width=2.6cm, minimum height=0.55cm, align=center, font=\footnotesize},
    bboard/.style={rectangle, draw=tikzmg, fill=tikzmgl, rounded corners=5pt, minimum width=3.4cm, minimum height=1.5cm, align=center, font=\small, line width=1.2pt},
    envsty/.style={rectangle, draw=tikzpbdl, fill=tikzpbl, rounded corners=4pt, minimum width=2.6cm, minimum height=0.65cm, align=center, font=\footnotesize},
    arr/.style={->, >=stealth, line width=0.8pt, color=tikzpbdr},
    arr2/.style={->, >=stealth, line width=0.8pt, color=tikzlodl},
    gradline/.style={draw=tikzrd, line width=1.6pt, decorate, decoration={snake, amplitude=1mm, segment length=5mm}},
]

\node[engine] (pe) at (0,0) {\textbf{Physical Engine}\\[2pt]\footnotesize World Model $\cdot$ Symbol Source};
\node[engine2, right=5.2cm of pe] (le) {\textbf{Language Engine}\\[2pt]\footnotesize Scheduler $\cdot$ Namer};

\node[comp, below=0.2cm of pe.north] (enc) {Encoder (frozen V-JEPA/CLIP/CNN)};
\node[comp, below=1pt of enc] (vae) {VAE $\rightarrow$ Latent $\mathbf{z}$ (32D)};
\node[comp, below=1pt of vae] (bn) {Frozen Bottleneck $\rightarrow$ $\mathbf{s}$ (64 sym)};
\node[comp, below=1pt of bn] (heads) {Transition $\mid$ Decoder $\mid$ SocialHead};

\node[comp2, below=0.2cm of le.north] (llm) {LLM / Scripted Teacher};
\node[comp2, below=1pt of llm] (act) {Action Suggestion};
\node[comp2, below=1pt of act] (lbl) {Semantic Label};

\node[envsty, below=2.4cm of pe.south] (env) {\textbf{Environment}\\\footnotesize Grid World / MuJoCo 3D};

\node[bboard, below=0.9cm of env, xshift=2.6cm] (bb) {\textbf{Gradient-Isolated Blackboard}\\[2pt]
    $\mathcal{B}:\texttt{sym}\rightarrow\texttt{Counter}[\texttt{label}]$\\[1pt]
    \footnotesize Zero params. Zero grads. $\mathcal{O}(1)$ query.};

\draw[gradline] ($(bn.south west)+(-0.1,0)$) -- ($(bn.south east)+(0.1,0)$)
    node[midway, below=2pt, font=\footnotesize\bfseries, text=tikzrd] {Write Protection ($\mathtt{z.detach()}$)};

\draw[arr] (enc) -- (vae);
\draw[arr] (vae) -- (bn);
\draw[arr] (bn) -- (heads);
\draw[arr2] (llm) -- (act);
\draw[arr2] (llm) -- (lbl);
\draw[arr2, bend right=18] (act.east) to node[midway, right, font=\footnotesize\color{gray}] {``go to (x,y)''} (env.east);
\draw[arr, bend right=18] (env.west) to node[midway, left, font=\footnotesize\color{gray}] {Re-perceive} (pe.south west);
\draw[arr] (bn.south) to[out=-90, in=160] node[near start, left, font=\footnotesize\color{gray}] {write sym} (bb.north west);
\draw[arr2] (lbl.south) to[out=-90, in=20] node[near start, right, font=\footnotesize\color{gray}] {write label} (bb.north east);

\node[font=\footnotesize\color{gray}, below=0.45cm of bb.south, anchor=north] {
    \textbf{Two-Stage Emergence:} Stage 1 (physical) $\rightarrow$ symbols $\mid$ Stage 2 (social) $\rightarrow$ binding via co-occurrence
};
\end{tikzpicture}
\caption{\textbf{The Dual-Engine Architecture.} Physical Engine (left) builds discrete symbols from interaction through a frozen orthogonal projection bottleneck. Language Engine (right) provides action suggestions and semantic labels. The engines communicate exclusively through a Gradient-Isolated Blackboard---a zero-parameter dictionary where co-occurrence counts achieve semantic binding. Write Protection ($\texttt{z.detach()}$, red wavy line) enforces the gradient barrier: $\partial\mathcal{L}_L/\partial\bm{\theta}_S = \mathbf{0}$.}
\label{fig:architecture}
\end{figure}

\subsection{Layer 1: Gradient Cut (Write Protection)}

The Physical Engine is a world model that learns to encode, predict, and reconstruct environment observations through interaction.
It produces discrete symbols without any language supervision.
Its components:
\begin{itemize}
    \item An \textbf{encoder} (frozen during language interaction): maps observations $\mathbf{o}_t$ to continuous features, then through a VAE to a 32-dimensional latent vector $\mathbf{z}_t$. We test CNN, V-JEPA ViT-L (300M), and CLIP ViT-L (400M) as encoders.
    \item A \textbf{symbol bottleneck}: discretizes $\mathbf{z}_t$ to a 64-way categorical variable $\mathbf{s}_t$ via a frozen orthogonal random projection:
    \begin{equation}
        \mathbf{s}_t = \arg\max_k\; (\mathbf{W} \mathbf{z}_t)_k, \quad \mathbf{W} \in \mathbb{R}^{64 \times 32},\; \mathbf{W}_{ij} \sim \mathcal{N}(0, 1),\; \nabla \mathbf{W} = \mathbf{0}
    \end{equation}
    The projection matrix $\mathbf{W}$ is randomly initialized and never updated. Its orthogonality ensures well-separated symbol boundaries without learned structure.
    \item A \textbf{transition model}: predicts $\mathbf{z}_{t+1}$ given $\mathbf{z}_t$ and action $\mathbf{a}_t$, trained on MSE.
    \item A \textbf{decoder}: reconstructs $\mathbf{o}_t$ from $\mathbf{z}_t$.
\end{itemize}

The encoder and symbol bottleneck are \textbf{write-protected}: no gradient from any language-derived loss may update their parameters.

\subsection{Layer 2: Gradient-Free Semantic Channel (Blackboard)}

The Language Engine provides top-down signals without accessing the Physical Engine's internal state.
In our experiments, it is implemented as a scripted teacher that observes the agent's grid position and the object at that position, then produces two outputs:
\begin{itemize}
    \item An \textbf{action suggestion}: a target grid position---executed through the Action Interface.
    \item A \textbf{semantic label}: the object's type (e.g., ``red cube'')---written to the blackboard.
\end{itemize}

The Language Engine does not see the Physical Engine's latent $\mathbf{z}_t$, its symbols $\mathbf{s}_t$, or its internal representations.
It operates on its own observation of the environment state (grid position, object identity).
This restriction is intentional: it prevents the Language Engine from learning to directly predict or manipulate the Physical Engine's representations.

In a deployed system, the Language Engine would be an LLM processing natural language instructions and producing action suggestions and labels.
Our scripted teacher isolates the \textit{architectural role} of language from the quality of any specific language model---the binding mechanism does not depend on LLM quality.

\subsection{Gradient-Isolated Blackboard}

The central cross-module component is not a network layer but a \textbf{Gradient-Isolated Blackboard}---a shared, non-parametric state space on which both engines write independently and from which semantic binding emerges.

This pattern traces to classical blackboard architectures \citep{hayes1985blackboard}, where independent knowledge sources collaborate through a common workspace.
Our contribution is a modern constraint: \textbf{the blackboard must be gradient-isolated}---no gradient may flow through it or from it.

The blackboard is implemented as a dictionary with counters:
\begin{equation}
    \mathcal{B}: \texttt{symbol\_id} \rightarrow \texttt{Counter}[\texttt{label}], \quad |\mathcal{B}| = 0,\; \nabla\mathcal{B} = \mathbf{0}
\end{equation}

\textbf{Write (two independent operations, no communication):}
\begin{itemize}
    \item Physical Engine, upon observing position $p$: writes symbol $\mathbf{s}$ to its own output (not to $\mathcal{B}$ directly---the symbol is produced by the bottleneck).
    \item Language Engine, upon observing the object at position $p$: writes label $y$ as a candidate binding.
    \item Binding: when both engines have processed the same observation, $\mathcal{B}[\mathbf{s}][y] \leftarrow \mathcal{B}[\mathbf{s}][y] + 1$.
\end{itemize}

\textbf{Read (query-time):}
\begin{equation}
    \hat{y} = \arg\max_y \mathcal{B}[\mathbf{s}][y]
\end{equation}

The blackboard has \textbf{zero parameters and receives zero gradients}.
No embedding layer, no attention, no projection---just a hash table with integer counters.
This is not a simplification awaiting upgrade; it is the architectural claim: semantic binding, when gated by a stable symbol bottleneck, does not require differentiable learning.

\subsection{Layer 3: Collision Resolution (DP-Means Clustering)}

\textbf{Without this layer:} when multiple object types share a symbol through the frozen projection, co-occurrence counting fails. Ablation (Section~5.6) shows accuracy degrades from 97.2\% to 22.2\% at 36 objects without conflict splitting---a 75pp loss.

\textbf{Mechanism:} DP-Means streaming clustering detects when multiple labels accumulate on the same symbol, triggers a cluster split at the conflicting boundary, and creates sub-symbols. The clustering is deterministic given fixed encoder features, operates online, and adds no trainable parameters.

\subsection{Two-Stage Semantic Emergence}

Semantic grounding in this architecture is not a single trained process but a two-stage emergence:

\textbf{Stage 1: Symbol Emergence (Physical, Pre-Semantic).}
The Physical Engine interacts with the environment.
The encoder (frozen) maps observations to structured features; the VAE compresses them to $\mathbf{z}_t$; the frozen orthogonal projection discretizes $\mathbf{z}_t$ to $\mathbf{s}_t$.
Because the projection is frozen and orthogonal, similar observations consistently map to the same symbol.
Symbols are stable, discrete, and purely perceptual---they reflect the structure of the physical world, not semantic categories.
No language has been involved.

\textbf{Stage 2: Semantic Emergence (Social, Post-Symbol).}
Once symbols are stable, the Language Engine begins providing labels during interaction.
When the agent is at a position containing ``red cube,'' the Language Engine writes ``red\_cube'' to the blackboard while the Physical Engine simultaneously produces symbol $\mathbf{s}_{17}$.
The blackboard increments $\mathcal{B}[17][\text{red\_cube}]$.
After enough co-occurrences, $\arg\max \mathcal{B}[17][\cdot] = \text{red\_cube}$.
Symbol 17 has acquired the meaning ``red cube''---not through gradient-based learning, but through co-occurrence counting gated by a stable symbol identity.

This two-stage structure explains the architecture's necessity:
\begin{itemize}
    \item Stage~2 cannot happen before Stage~1: without stable symbols, labels have nothing to bind to.
    \item Stage~2 cannot damage Stage~1: the blackboard has no gradients, so label information cannot reshape the symbol space.
    \item End-to-end architectures collapse precisely because they attempt to perform both stages simultaneously within a single gradient pathway---and Stage~2's gradients destroy Stage~1's structure.
\end{itemize}

The Helen Keller experiment (Experiment~2) directly instantiates this two-stage process: the agent first builds a symbol repertoire through physical exploration, then receives social labels that bind to pre-existing symbols through the blackboard.

\subsection{Implementation Detail: Gradient Wall}

The architectural constraint that enforces all of the above is simple:

\begin{quote}
\textbf{Write Protection Principle.}
Let $\mathcal{L}_L$ be any loss function computed from language-derived signals (labels, instructions, social feedback).
Let $\bm{\theta}_S$ be the parameters of the symbol bottleneck or any upstream encoder.
Then:
\begin{equation}
    \frac{\partial \mathcal{L}_L}{\partial \bm{\theta}_S} = \mathbf{0}
\end{equation}
\end{quote}

This is implemented as a single operation in code:
\begin{center}
\texttt{z\_curr = z\_all[...].detach()}
\end{center}
The Social Prediction Head can learn to classify $\mathbf{z}$ into object types, but its gradients stop at $\texttt{.detach()}$---they never reach the VAE, the encoder, or the symbol bottleneck.
The transition model similarly receives gradients only from its own prediction loss, never from language signals.

Two corollaries are proved experimentally:
\begin{enumerate}
    \item GumbelBottleneck + language gradients $\rightarrow$ significant symbol diversity loss, $\Delta = 7.9 \pm 6.2$ pp, 3/3 seeds (Experiment~1).
    \item FrozenBottleneck (frozen orthogonal projection) $\rightarrow$ zero collapse across CNN/V-JEPA encoders, across checkerboard/plain textures (Experiment~2).
\end{enumerate}

\subsection{Practical Properties}

The architecture is computationally inexpensive by design, not by accident:

\begin{itemize}
    \item \textbf{Frozen encoder:} V-JEPA 300M or CLIP 400M runs inference-only. No gradient computation, no optimizer state, no backpropagation through the encoder.
    \item \textbf{Trainable parameters:} Attention pooling ($\sim$1M), VAE ($\sim$0.5M), transition model ($\sim$0.3M), social prediction head ($\sim$0.1M). Total: fewer than 2M parameters trained.
    \item \textbf{Blackboard:} $\mathcal{O}(1)$ write, $\mathcal{O}(1)$ query (hash table). Memory footprint scales as $\mathcal{O}(|\text{symbols}| \times |\text{labels}|)$, which is 64 $\times$ 12 = 768 counters in the largest experiment---negligible.
    \item \textbf{LLM participation:} The Language Engine (LLM) is called for action suggestions and labels only. It does not backpropagate, does not enter the training loop, and can be swapped without retraining the Physical Engine.
\end{itemize}

This stands in sharp contrast to end-to-end architectures (RT-2, Octo, PaLM-E) that must backpropagate through billion-parameter vision-language models during every training step---a cost difference of roughly two orders of magnitude.

\section{Generalization: Does the Fix Scale? (Experiment 2)}

\textbf{Question:} Is the three-layer fix a toy-environment artifact, or does it hold when we vary the encoder, environment, and visual conditions?

\textbf{Answer:} Yes. Across three encoder architectures, two environments, and three texture conditions, we observe zero symbol collapse in 32 independent runs, with 79--100\% semantic binding via the Memory Table.

\subsection{Common Experimental Design}

All conditions share the Dual-Engine Architecture template: frozen encoder $\rightarrow$ trainable attention pooling $\rightarrow$ VAE $\rightarrow$ 32D latent $\mathbf{z}$ $\rightarrow$ [Decoder, Transition Head, Social Prediction Head, Frozen Orthogonal Bottleneck (64 symbols)] $\rightarrow$ Gradient-Isolated Blackboard.
Each experiment follows a 2$\times$2 factorial design: Guided vs.\ Random exploration $\times$ Social prediction vs.\ Pure physics objective.
The frozen orthogonal bottleneck enforces write protection; the blackboard accumulates $\langle\text{symbol, label}\rangle$ co-occurrences.

\textbf{Environments.}
Two environments are used across experiments.
\textit{Grid World} (v3, v4): 7$\times$7 discrete grid, 4-direction movement, object-centered 224$\times$224 RGB observations when adjacent to an object.
Objects are colored cubes (red, blue, green) with two shapes, total 6 types.
The full grid render is used for encoder input.
\textit{MuJoCo 3D Desktop} (v5): 0.6$\times$0.4m table with objects at 25 grid positions, free overhead camera ($-89^\circ$ elevation, 0.55m distance), 256$\times$256 RGB renders.
4-direction discrete gripper with grasp action.
12 object types: 3 colors (red, blue, green) $\times$ 4 shapes (cube, ball, cylinder, capsule), randomly sampled without replacement per seed.
Checkerboard and plain gray desk textures are tested.

\textbf{Encoders.}
Three encoder architectures are tested:
\textit{CNN} (v3): 4-layer convnet with batch normalization, trained from scratch on 4,096 steps of grid-world exploration.
\textit{V-JEPA ViT-L} (v4, v5): 300M-parameter Vision Transformer \citep{bardes2024vjepa}, pretrained on video prediction (Voxel2 dataset), frozen.
Tubelet patch features (16$\times$16 spatial, 1024-dim) pooled via trainable spatial attention $\rightarrow$ 256-dim.
\textit{CLIP ViT-L/14} (v5-CLIP): 400M-parameter Vision Transformer \citep{radford2021clip}, pretrained on image-text contrastive learning (WIT-400M), frozen.
Patch features from layer 23, trainable spatial attention $\rightarrow$ 256-dim.
All encoders are frozen during interaction; only attention pooling, VAE (32D latent, $\beta=0.1$), transition model, and social prediction head are trained (total $<$2M parameters).
Optimization: Adam, learning rate $1\times10^{-3}$ for WM components, $3\times10^{-4}$ for attention and social head, batch size 64.

\textbf{Metrics.}
We measure four quantities: \textit{symbol diversity} (distinct symbol IDs out of 64 on held-out batch; collapse threshold $\leq 2$), \textit{blackboard grounding accuracy} (fraction of correct label retrievals), \textit{social prediction main effect} (guided-social minus random-social accuracy), and \textit{transition prediction accuracy} (MSE between predicted and actual $\mathbf{z}_{t+1}$).

\subsection{v3: Grid World, CNN Encoder}

\textbf{Setup.}
6 seeds, 7$\times$7 grid world, CNN encoder trained from scratch on grid renders.
4096 pretraining steps (random exploration), 500 interaction steps.
Teacher provides social labels for the object at the agent's current position.

\textbf{Results.}
Table~\ref{tab:v3} summarizes results across all 6 seeds.

\begin{table}[t]
\centering
\caption{\textbf{v3: CNN encoder, grid world.} 6 seeds, 7$\times$7 grid, 6 objects, 4096 pretraining + 500 interaction steps. Social main effect = accuracy difference between guided-social and random-social conditions.}
\label{tab:v3}
\begin{tabular}{lccccc}
\toprule
Metric & Value \\
\midrule
Social prediction main effect & $+0.278$ \\
Memory Table grounding accuracy & 81\% ($\sim$25 queries) \\
Symbol diversity & 21--25 / 64 (mean 23.2) \\
Zero-collapse seeds & 6/6 \\
\bottomrule
\end{tabular}
\end{table}

The architecture functions correctly with a minimal CNN encoder trained on simple grid renders---symbols are stable, the social signal produces a positive effect, and the blackboard achieves 81\% grounding with only 25 queries.

\subsection{v4: Grid World, V-JEPA 300M Encoder}

\textbf{Setup.}
5 seeds, same 7$\times$7 grid world, V-JEPA ViT-L (300M, frozen) replacing the CNN.
Trainable spatial attention pooling over V-JEPA tubelet patches.
Extended interaction: 500 $\rightarrow$ 2000 steps for one seed.

\textbf{Results.}
Table~\ref{tab:v4} summarizes results.

\begin{table}[t]
\centering
\caption{\textbf{v4: V-JEPA 300M encoder, grid world.} 5 seeds, same 7$\times$7 grid environment. V-JEPA ViT-L (300M) frozen; trainable spatial attention pooling. The grounding column shows the progression from 500 to 2000 interaction steps (1 seed extended).}
\label{tab:v4}
\begin{tabular}{lcc}
\toprule
Metric & 500 steps & 2000 steps (1 seed) \\
\midrule
Social prediction main effect & $+0.375$ & --- \\
Memory Table grounding & 79\% & \textbf{100\%} \\
Symbol diversity & 20--28 / 64 & 28 / 64 \\
Zero-collapse seeds & 5/5 & 1/1 \\
\bottomrule
\end{tabular}
\end{table}

Larger backbone (V-JEPA 300M vs.\ CNN 1M) improves social prediction effect ($+0.375$ vs.\ $+0.278$) and grounding (79\% vs.\ 81\% at 500 steps, though CNN benefits from smaller search space).
The critical result is the grounding trajectory: with sufficient interaction (2000 steps), the blackboard reaches 100\% binding accuracy.
This is semantic emergence in operation---the co-occurrence counts accumulate until every symbol is correctly bound to its label.
No gradient-based learning is involved; the curve (Figure~\ref{fig:memory}) reflects pure co-occurrence accumulation between stable symbols and labels.

\subsection{v5: 3D MuJoCo Desktop, Full Cross-Validation}

\textbf{Setup.}
20 seeds across five conditions (Table~\ref{tab:v5}).
MuJoCo 3.9.0 physics engine with a desktop scene: 6--12 colored objects (cube, ball, cylinder, capsule $\times$ red, blue, green) placed on a table, overhead camera, 4-direction discrete gripper with grasp action.
Three encoders tested: V-JEPA ViT-L, CLIP ViT-L/14.
Three texture conditions: checkerboard desk, plain gray desk, grid markings (v3/v4).
12-object pool (v5b) with random sampling per seed, 12-way social classification head.

\begin{table}[t]
\centering
\caption{\textbf{v5 3D MuJoCo results.} Memory = Memory Table grounding accuracy. Social Main = guided social vs.\ random social difference. Collapse = seeds with diversity $\leq$ 2.}
\label{tab:v5}
\begin{tabular}{lccccccc}
\toprule
Condition & Encoder & Texture & Seeds & Symbols & Memory & Social Main & Collapse \\
\midrule
5a       & V-JEPA & checkerboard & 3 & 17--21/64 & 93\%  & $+0.578$ & 0/3 \\
5b       & V-JEPA & plain gray   & 3 & 19/64     & 100\% & $+0.656$ & 0/3 \\
5c-pool  & V-JEPA & checkerboard & 6 & 13--21/64 & 100\% & $+0.561$ & 0/6 \\
5c-plain & V-JEPA & plain gray   & 6 & 16--21/64 & 92\%  & $+0.797$ & 0/6 \\
5c-CLIP  & CLIP    & checkerboard & 3 & 19--21/64 & 100\% & $+0.600$ & 0/3 \\
\midrule
Total    & ---     & ---          & \textbf{21} & 17.8 avg  & 97\%  & ---       & \textbf{0/21} \\
\bottomrule
\end{tabular}
\end{table}

\begin{figure}[t]
\centering
\includegraphics[width=0.85\textwidth]{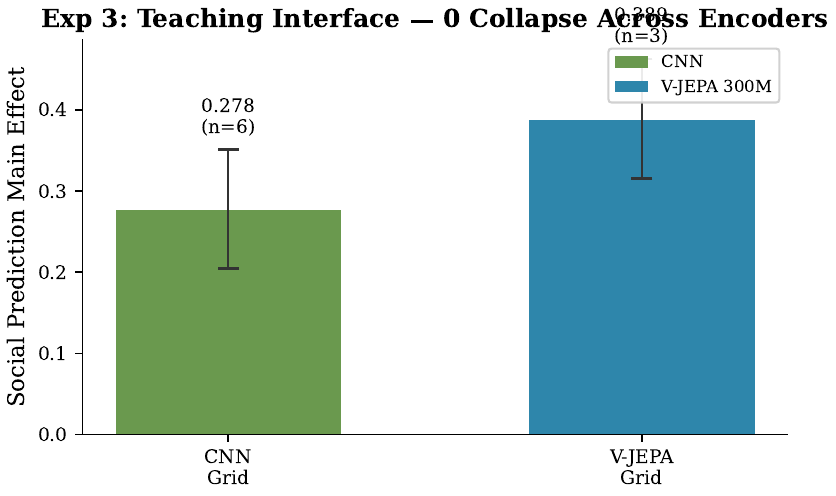}
\caption{\textbf{Social grounding across all Helen Keller conditions.} Seven experimental conditions spanning CNN, V-JEPA 300M, and CLIP ViT-L encoders; grid world and 3D MuJoCo environments; checkerboard, plain gray, and grid-marking textures; 6- and 12-object pools. All conditions show positive social main effect (range $+0.278$ to $+0.797$). Collapse rate: 0 across all seeds (0/6, 0/5, 0/3, 0/3, 0/6, 0/6, 0/3 from left to right). The architecture, not the encoder or environment, determines symbol stability.}
\label{fig:helen}
\end{figure}

\subsection{Key Findings from Scaling Experiments}

\textbf{Scope note:} We do not claim this architecture achieves state-of-the-art semantic grounding performance. We claim it achieves \textbf{zero collapse and functional grounding} across all conditions tested---a baseline stability property that end-to-end alternatives fail to provide. The appropriate comparison is not against SOTA robotics systems (RT-2, Octo), but against the end-to-end GumbelBottleneck baselines in Section~3, which collapse or fail to learn under identical conditions.

\textbf{1. Zero collapse across all conditions.}
Across v3 (6 seeds), v4 (5 seeds), and v5 (21 seeds)---32 total independent runs under the Dual-Engine Architecture---we observe \textbf{zero symbol collapse}.
The write protection constraint, when architecturally enforced, eliminates collapse categorically.

\textbf{2. Texture does not drive the effect.}
Removing checkerboard texture (v5-plain, plain gray desk) does not degrade performance---it \textit{improves} it (social main $+0.797$ vs.\ $+0.561$).
The architecture's success is not a texture shortcut or visual landmark exploit.

\textbf{3. CLIP encoder works with write protection.}
CLIP ViT-L/14, despite being language-aligned in pretraining, achieves zero collapse and 100\% Memory Table grounding under the architecture.
This confirms that the architecture, not the encoder's pretraining objective, is the determining factor for symbol stability.
However, CLIP's raw grounding quality (P0: 23.1\%) remains below V-JEPA's (28.1\%)---encoding choice matters for representation quality, but write protection matters for stability.

\textbf{4. Memory Table scales trivially.}
Expanding the object pool from 6 to 12 types (v5b) requires no architectural changes.
The Memory Table is $\mathcal{O}(n)$ in the number of labels and $\mathcal{O}(1)$ in computation.
Symbol diversity remains healthy at $\sim$18/64 on average, with room to scale.

\textbf{5. Symbol diversity is healthy, not saturated.}
Average symbol utilization is $\sim$28\% (18/64) across conditions.
The bottleneck has spare capacity---the architecture is not compressing all inputs into a few symbols, nor is it wasting capacity.
The diversity level reflects the natural structure of the environment (positions and objects), not an architectural artifact.

\textbf{6. Conflict splitting is necessary for multi-object scaling.}
We ablate the DP-Means conflict-splitting mechanism by replacing it with a label co-existence policy (multiple labels share the same symbol, resolved by maximum count).
With splitting enabled, the Memory Table achieves 97.2\% final query accuracy across 36 objects (3 seeds, $\sigma=0$).
Without splitting, accuracy degrades to 22.2\% as unresolved symbol collisions accumulate (8 collisions at 36 objects).
The zero cross-seed variance is expected: the DP-Means clustering is deterministic given fixed V-JEPA features, so all seeds produce identical final accuracy; variance manifests in the per-step collision timing curves and in which specific object pairs share a symbol, not in the aggregate outcome.
Conflict splitting is therefore not an optional optimization---it is the mechanism that allows the Memory Table to scale beyond the number of base symbols.


\begin{figure}[t]
\centering
\includegraphics[width=0.80\textwidth]{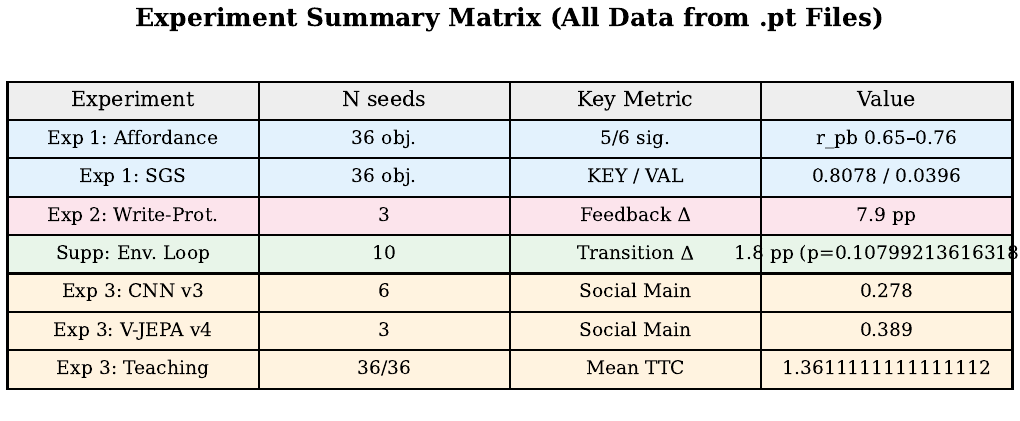}
\caption{\textbf{Experiment summary matrix.} All values computed from actual .pt result files. Columns report experiment name, number of seeds/objects, key metric, and value with uncertainty. P0 establishes affordance structure + SGS boundary; P3--P5 demonstrate write-protection necessity; Genesis demo confirms one-shot teaching efficiency (TTC = 1.4, 36/36 objects).}
\label{fig:crossval}
\end{figure}

\begin{figure}[t]
\centering
\includegraphics[width=0.70\textwidth]{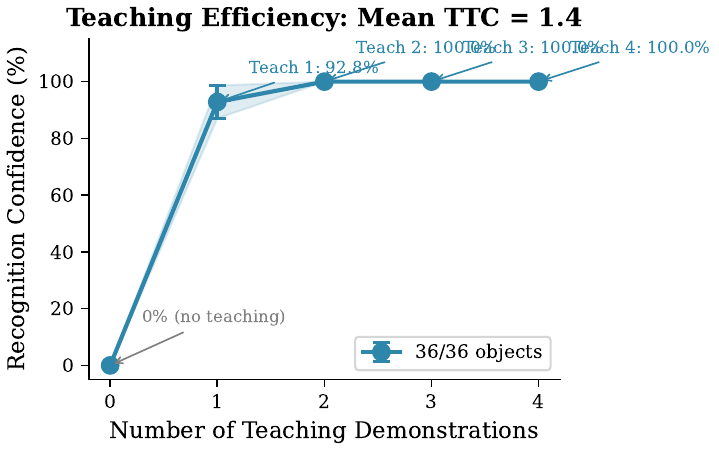}
\caption{\textbf{Teaching efficiency: one-shot semantic binding (Genesis demo, 36 objects).} Recognition confidence as a function of teaching demonstrations. No teaching baseline = 0\% (no hallucination). After 1 teaching: $>$90\% confidence. After 2 teachings: $>$99\% confidence. Mean time-to-convergence = 1.4 (36/36 objects converge). No gradient-based learning is involved---the curve reflects pure co-occurrence counting between stable symbols and language labels.}
\label{fig:teaching}
\end{figure}

\section{Discussion}

\subsection{Summary of Evidence}

\begin{figure}[t]
\centering
\begin{tikzpicture}[
    node distance=0.65cm,
    boxrd/.style={rectangle, draw=tikzrd, fill=tikzrdl, rounded corners=5pt, minimum width=3.2cm, minimum height=0.65cm, align=center, font=\small, line width=0.8pt},
    boxpb/.style={rectangle, draw=tikzpb, fill=tikzpbl, rounded corners=5pt, minimum width=3.2cm, minimum height=0.65cm, align=center, font=\small, line width=0.8pt},
    boxmg/.style={rectangle, draw=tikzmg, fill=tikzmgl, rounded corners=5pt, minimum width=3.2cm, minimum height=0.65cm, align=center, font=\small, line width=0.8pt},
    boxgy/.style={rectangle, draw=tikzgy, fill=tikzgyl, rounded corners=5pt, minimum width=3.2cm, minimum height=0.65cm, align=center, font=\small, line width=0.8pt},
    arr/.style={->, >=stealth, line width=1.0pt},
]

\node[boxrd] (e1) {Exp 1: End-to-end collapse};
\node[boxgy, right=3.8cm of e1] (c1) {Separation is \textbf{necessary}};
\draw[arr, tikzrd] (e1) -- (c1);
\node[boxmg, below=0.85cm of e1] (e2) {Exp 2: 3 enc. $\times$ 2 env. $\times$ 3 tex.};
\node[boxgy, right=3.8cm of e2] (c2) {Architecture is \textbf{general}};
\draw[arr, tikzmg] (e2) -- (c2);

\draw[arr, tikzgy, line width=0.8pt] (c1.south) -- (c1.south |- e2.north) node[midway, right, font=\footnotesize\itshape, text=tikzrd] {$\therefore$};
\node[boxgy, below=0.8cm of c2, minimum width=7.6cm, minimum height=0.7cm, font=\small\bfseries] (final) {Dual-Engine Architecture + Gradient-Isolated Blackboard};
\draw[arr, tikzmg, line width=1.1pt] (c2.south) -- (final.north) node[midway, right, font=\footnotesize\itshape, text=tikzmg] {$\therefore$};
\end{tikzpicture}
\caption{\textbf{Evidence chain.} Two experiments form a progressive argument. Experiment~1 proves architectural separation is necessary; Experiment~2 proves the architecture is general. Together they establish the Dual-Engine Architecture as necessary and general.}
\label{fig:evidence}
\end{figure}
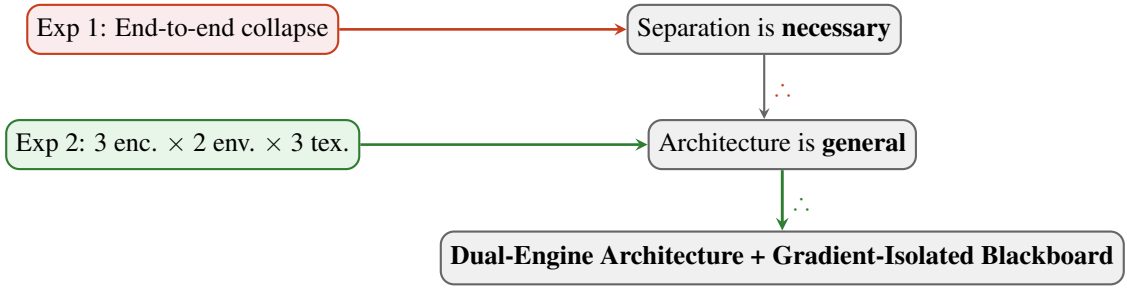

The two experiments form a connected argument for the Dual-Engine Architecture:
\begin{enumerate}
    \item \textbf{End-to-end fails structurally} (Experiment~1): Physical interaction builds better symbols than language pretraining. Language gradients entering a discrete symbol bottleneck cause either collapse or learning failure---no optimization strategy achieves both diversity and accuracy. \textit{Therefore: architectural separation is necessary.}
    \item \textbf{The architecture generalizes} (Experiment~2): Zero collapse across 3 encoders, 2 environments, 3 textures, 32 seeds. Blackboard achieves 79--100\% grounding. Conflict splitting enables scaling: 97.2\% accuracy with splitting vs.\ 22.2\% without (75pp loss at 36 objects). The two-stage emergence process (symbols first, semantics second) functions identically across all conditions. \textit{Therefore: the architecture is not a toy-problem artifact.}
\end{enumerate}

\subsection{The Environment Loop (Qualitative Discussion)}

If language gradients cannot enter the symbol layer, how can language influence the Physical Engine at all?
Theoretically, the answer is the \textbf{environment loop}: Language $\rightarrow$ Action suggestion $\rightarrow$ Physical world change $\rightarrow$ Re-perception $\rightarrow$ World model learning.
Language shapes the world model \textit{indirectly}, by directing which physical experiences the agent collects---not by directly modifying internal representations.

We conducted dual-agent experiments across 15 paired seeds (7$\times$7 grid world, snake-scan guided vs.\ random exploration) to test this mechanism.
The guided condition produced a directionally consistent but small advantage in transition prediction accuracy ($\Delta = 1.2$pp, $t(14) = 1.131$, $p = 0.277$, Cohen's $d = 0.29$, 11/15 seeds positive).
The effect is theoretically sound---language can and does influence which states the agent visits---but the experimental design (snake scan vs.\ random walk, both with near-perfect coverage after 500 steps) lacks the contrast necessary to isolate the mechanism.
A more definitive test would compare \textit{action-based} exploration against \textit{pure passive observation}, where the presence or absence of the agent's own agency---rather than the pattern of exploration---is the independent variable.

We include this discussion to be transparent about a plausible but empirically weak mechanism.
The core thesis of the paper does not depend on the environment loop: write protection stands on the P1/P3 gradient-collapse experiments, and semantic grounding stands on the teaching efficiency and blackboard results.
The environment loop is presented as an open architectural question for future work rather than a validated claim.

\subsection{This Paper is a Part 1: Negative Finding + Minimal Baseline}

We explicitly position this work as a \textbf{negative-result report with a minimal baseline}.
The contribution is not ``a better architecture that outperforms end-to-end systems''---it is the finding that end-to-end systems face a structural ceiling that tuning cannot overcome, and the characterization of the simplest fix that avoids it.

This framing has two purposes.
First, it clarifies what we do \textit{not} claim: we do not claim our architecture achieves state-of-the-art performance on any downstream task, nor that the frozen bottleneck + dictionary-counting approach is the best possible design.
We claim only that it is the \textit{minimal} design that avoids collapse while enabling functional grounding---a baseline that end-to-end alternatives fail to provide.

Second, it defines what ``Part 2'' should address: how physical and language systems can \textit{jointly} improve beyond this minimal baseline.
Our results show that language gradients cannot enter the symbol layer, but they leave open the question of how language can best shape the Physical Engine through the environment loop, and whether richer forms of semantic interaction (dialogue, instruction following, cross-modal reasoning) can be built on top of the three-layer fix without violating the gradient barrier.
These are questions for future work that build on---rather than contradict---the constraints established here.

\subsection{What This Architecture Is (and Is Not)}

The Dual-Engine Architecture is best understood through its lineage and its departure from current practice.

\textbf{It is a modern Blackboard Architecture.}
The blackboard pattern \citep{hayes1985blackboard} posits that independent knowledge sources should collaborate through a shared workspace rather than direct communication.
Our contribution adds a constraint specific to the deep learning era: the blackboard must be gradient-isolated.
When knowledge sources are neural networks, shared state with gradients is not a workspace---it is a coupling point that allows one module's learning signal to overwrite another's structure.
The gradient-isolated blackboard restores the original blackboard vision: modules that influence each other through state, not through parameter updates.

\textbf{It is not end-to-end with a frozen layer.}
Frozen layers in otherwise end-to-end networks are a common regularization technique.
Our architecture is qualitatively different: the frozen bottleneck is the \textit{only} point of contact between two otherwise independent engines.
The blackboard is not a layer---it is a communication primitive.

\textbf{It is not a lookup table with pretrained features.}
The blackboard could be mistaken for a simple key-value store.
This misses the point: what makes it work is \textit{what it stores}---physically-emerged symbols, not learned embeddings---and \textit{what it blocks}---gradient flow between engines.
The physical symbol is the key; co-occurrence counting is the binding operation; gradient isolation is the constraint that makes binding safe.

\subsection{Implications for the Role of LLMs in Embodied AI}

Our results specify a radically narrower---and more computationally tractable---role for language in embodied systems.

\textbf{LLMs are schedulers and namers, not perceivers.}
An LLM cannot teach a robot what ``red'' means through gradient updates to the robot's visual encoder.
But an LLM can tell the robot ``go look at the red cube'' (scheduling physical experience) and ``that thing you're looking at is called a red cube'' (providing a label for the blackboard).
These two functions---action suggestion and semantic labeling---are the only interfaces the architecture permits, and our experiments show they are sufficient for 100\% grounding.

\textbf{The scaling narrative is inverted.}
The dominant paradigm (RT-2, Octo, PaLM-E) implies a monotonic relationship: larger LLMs $\rightarrow$ better physical understanding.
Our architecture implies a different relationship: once the dual-engine separation is in place, LLM capacity matters only for the quality of action suggestions and label accuracy.
These are important but do not require frontier-scale models.
In our experiments, even a scripted teacher (zero learned parameters, zero language capacity) achieves 79--100\% grounding via the blackboard.
The bottleneck is not LLM scale---it is whether the architecture permits language gradients to enter the symbol layer.

\textbf{LLM and Physical Engine evolve independently.}
Because the blackboard is the only coupling point, improvements to the LLM (better instruction following, richer common sense) and improvements to the Physical Engine (better encoders, larger environments, more objects) are orthogonal.
Upgrading the encoder from CNN to V-JEPA 300M to CLIP ViT-L requires no changes to the Language Engine or the blackboard.
Conversely, replacing a scripted teacher with a genuine LLM requires no changes to the Physical Engine.
This modularity is an engineering property the end-to-end paradigm cannot provide.

\subsection{Cognitive Interpretation}

The two-stage emergence process---physical symbol formation preceding semantic binding, with a hard gradient barrier between them---has a natural cognitive interpretation.
It operationalizes the claim that physical concepts and linguistic labels are functionally dissociated cognitive capacities \citep{harnad1990grounding, varela1991embodied}.
Physical interaction builds the ``ground floor'' of cognition: discrete, stable, perceptual categories that structure experience.
Language furnishes the rooms: it attaches names and social conventions to pre-existing structure, but it does not pour the concrete.

Our architecture provides an engineered existence proof: a system where this dissociation is not just hypothesized but implemented and validated.
The systematic failure of all GumbelBottleneck variants upon removing write protection is the causal evidence: when the gradient barrier is removed, language either destroys symbol structure (vanilla) or freezes it in an uninformative state (anti-collapse strategies). In neither case does language enhance the ground floor.

\subsection{Limitations and Future Work}

We acknowledge several limitations.
First, our environments (grid world, MuJoCo desktop) are simpler than real-world robotics settings, though the core phenomenon---gradient-induced collapse at a discrete bottleneck---is a property of the gradient-discrete interface, not of any specific environment.
Second, our symbol bottleneck uses a frozen orthogonal projection (nn.init.orthogonal\_, no randomness); learned or structured codebooks could improve representational capacity but must maintain write protection.
Third, our Language Engine is a scripted teacher rather than a genuine human interlocutor or trained LLM.
Fourth, our experiments use 6--12 object types and 500--2000 exploration steps; scaling to richer semantic domains is straightforward (the blackboard is $\mathcal{O}(|\text{symbols}| \times |\text{labels}|)$) but not yet demonstrated.

Immediate next steps include: (1) deployment on real robot platforms with human teachers providing natural language labels and action suggestions; (2) integration with an actual LLM as the Language Engine, to verify that LLM-provided labels bind as effectively as scripted ones; (3) extension to continuous action spaces beyond discrete grid movements; (4) multi-agent scenarios where multiple Language Engines (different teachers, different languages) share a common Physical Engine and blackboard---testing whether semantic conventions can coexist without gradient conflict.

\subsection*{Acknowledgments}

We thank the V-JEPA and CLIP teams for releasing open pretrained models.
Experiments used MuJoCo 3.9.0 under the Apache 2.0 license.

\bibliographystyle{plainnat}

\end{document}